\documentclass{article}
\usepackage[colorlinks,linkcolor=red,hyperindex,bookmarks=false]{hyperref}
\usepackage{spconf,amsmath,graphicx}
\usepackage{multirow}

\title{Noise-Sampling Cross Entropy Loss: Improving Disparity Regression via Cost Volume Aware Regularizer}
\name{Yang Chen$^{1}$ \qquad Zongqing Lu$^{2}$ \qquad Xuechen Zhang$^{3}$\qquad Lei Chen$^{4}$ \qquad Qingmin Liao$^{5}$\thanks{Thanks to the Special Foundation for the Development of Strategic Emerging Industries of Shenzhen(JCYJ20170817161056260) for funding.}}

\address{Tsinghua University\\
Department of Electronic Engineering\\
Shenzhen 518055, China }
\begin{document}

\ninept

\maketitle

\begin{abstract}
Recent end-to-end deep neural networks for disparity regression have achieved the state-of-the-art performance. However, many well-acknowledged specific properties of disparity estimation are omitted in these deep learning algorithms. Especially, matching cost volume, one of the most important procedure, is treated as a normal intermediate feature for the following softargmin regression, lacking explicit constraints compared with those traditional algorithms. In this paper, inspired by previous canonical definition of cost volume, we propose the noise-sampling cross entropy loss function to regularize the cost volume produced by deep neural networks to be unimodal and coherent. Extensive experiments validate that the proposed noise-sampling cross entropy loss can not only help neural networks learn more informative cost volume, but also lead to better stereo matching performance compared with several representative algorithms.
\end{abstract}

\section{Introduction} \label{sec:intro}
Stereo matching is a classic but challenging problem in computer vision, aiming to offer a solution which recovers real-world 3D structure from 2D information, and it is widely used in various areas such as autonomous driving, augmented reality, remote sensing and robotic system. Given a pair of rectified stereo images, the stereo matching technique is to calculate the disparity, which measures the shift between one pixel and its counterpart pixel on the same horizon.

According to canonical taxonomy proposed by Schourstein \emph{et~al} \cite{scharstein2002taxonomy}, a stereo matching algorithm often consists of four steps: matching cost computation, cost aggregation, disparity regression and disparity refinement. Among these steps, matching cost computation, which aims to obtain cost volume, is regarded as the first and most important one \cite{scharstein2002taxonomy}. Generally speaking, cost volume is often a 3D tensor, where each element is defined to measure the matching cost between two points sampled from paired images but with the same horizon coordinate.  In traditional algorithms, early matching cost computation methods are often pixel-based ones \cite{anandan1989computational}. To improve the robustness and stability of the matching cost, many researchers proposed to compute the corresponding matching cost based on some sophisticated features \cite{zabih1994non,calonder2010brief}, and recent successful deep-learning methods inherit this wisdom. Zbontar \emph{et~al} \cite{zbontar2015computing} made the first step towards using deep convolutional nerual network (CNN) to learn deep representations for effective disparity calculation, achieving noticeable improvement compared with traditional methods. By adopting a more rational matching measure along with efficient network architecture, Content-CNN proposed by Luo \emph{et~al} \cite{luo2016efficient} also achieved impressive results. After obtaining deep representations, classical algorithms were used to get well-polished disparity. To sum up, these two pioneering deep learning methods differ from traditional feature-based ones by exploiting powerful representations learned by deep neural networks, while traditional algorithms mainly rely on handcraft features. Although deep features learned from deep models are  more representative, they are believed to lack interpretability compared with traditional ones. 

\begin{figure}[t]
\begin{minipage}[b]{1.0\linewidth}
  \centering
  \centerline{\includegraphics[width=7.5cm]{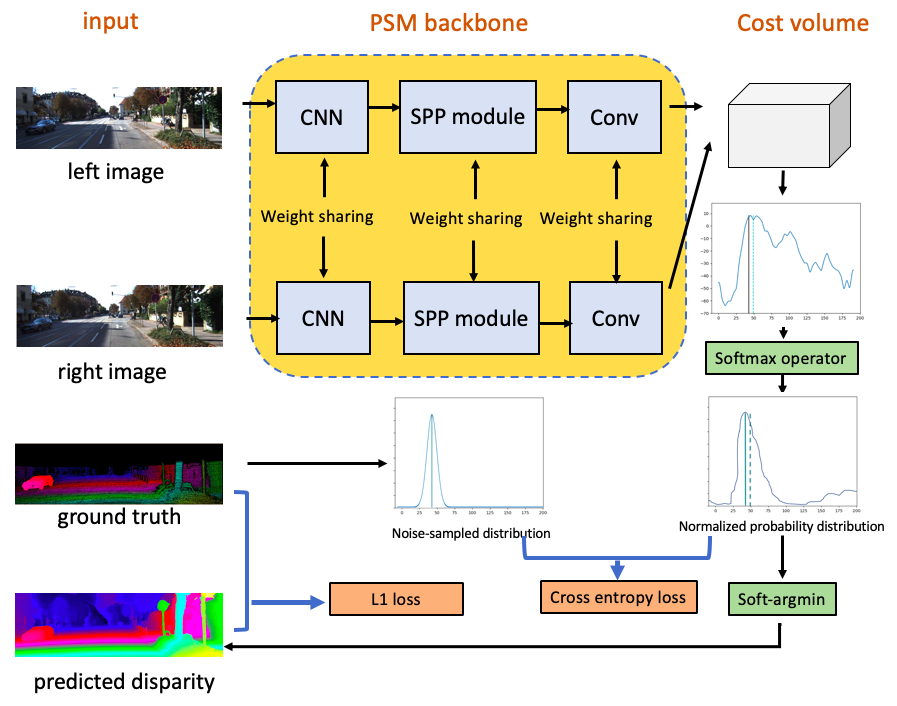}}
\end{minipage}
\caption{Schematic diagram of our method and network architecture. Our main contribution is proposing noise-sampling cross entropy losses to impose unimodality and coherence to the cost volume produced by deep disparity regression network.}
\label{fig:network}
\end{figure}

Along with the prevailing end-to-end deep learning framework, the state-of-the-art deep learning based stereo matching algorithms integrate the four steps above and train the overall network in an end-to-end way, where cost volume are often represented by certain intermediate features. To achieve end-to-end training of these deep models, Kendall \emph{et~at.} \cite{kendall2017end} proposed to use the differentiable soft-argmin operation as an alternative to the disparity regression step. With this method, the whole network can be optimized by regression loss and sub-pixel information can also be taken into account. Despite of the great progress made by soft-argmin operation, there are still two potential risks: one is that the cost volume constrained in this way may be in risk of multi modal distributed, meaning a relatively ``poor'' matching cost computation in the traditional sense; the other is cost volume of this kind may neglect the intrinsic coherence among the neighbor disparity. To preserve the advantages of the widely-used soft-argmin operation while mitigating its potential risks, inspired by one-hot cross entropy used in classification task \cite{krizhevsky2012imagenet} and regression models from Bayesian perspective \cite{murphy2012machine}, we propose the  noise-sampling cross entropy loss as an effective loss function term to address the mentioned issues.

Our contributions can be summarized from two aspects:

1. To help the network learn a more informative cost volume which is unimodal distributed and coherent, we propose the noise-sampling cross entropy loss as an effective loss function term used for deep disparity regression.

2. Experiments with the benchmarks demonstrate that the proposed noise-sampling cross entropy loss can improve the performance of soft-argmin based deep disparity estimation network. 

The remainder of this paper is organized as follows. In section \ref{sec: methods}, we explain our motivation and the rationality of the proposed loss function. In section \ref{sec:experiment}, experimental results comparedn with benchmarks and other methods are presented. We conclude the paper in section \ref{sec:concl}. 

\begin{figure*}[t]
\begin{minipage}[b]{1.0\linewidth}
  \centering
  \centerline{\includegraphics[width=18cm]{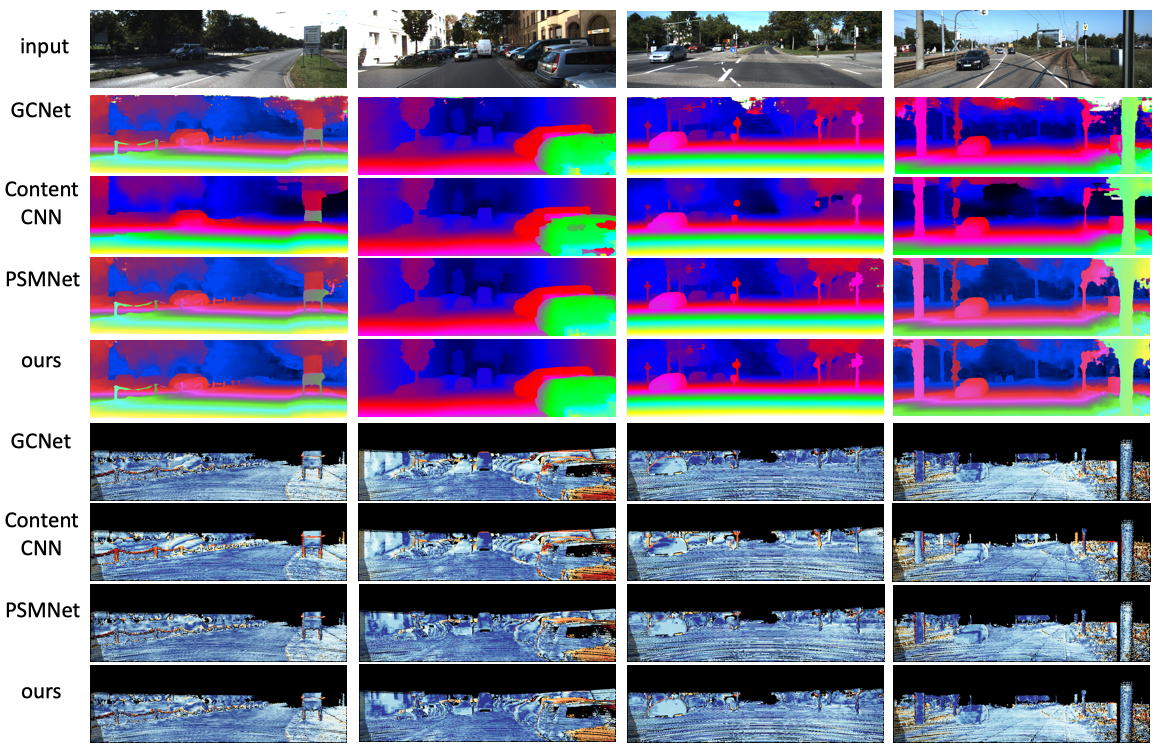}}
\end{minipage}
\caption{The results of disparity estimation on KITTI 2015 test images. The first row is the set of left input images. And the second to fifth rows are the output disparity estimations of GC-Net~\cite{kendall2017end}, Content-CNN~\cite{luo2016efficient}, PSMNet~\cite{chang2018pyramid}, noise-sampling PSMNet and in a top-down order. Accordingly, the sixth to ninth rows are the error maps in the same order.}
\label{fig:res}
\end{figure*}

\section{Motivation and Proposed methods} \label{sec: methods}
We begin by revisiting the softargmin operator in deep disparity regression. Based on powerful representations, disparity is often computed by a simple but effective strategy called ``winner-take-all" (WTA), where the disparity associated with the minimum cost value is chosen \cite{scharstein2002taxonomy}. However, this efficient opereator is not differentiable, which cannot be embedded into the modern end-to-end neural network directly. In GCNet, Kendall \emph{et~al} proposed to use softargmin along with regression loss such as mean square error (MSE) and mean absolute error (MAE) to address the problem. The softargmin operator in GCNet is defined as follows:
\begin{equation}
    \begin{split}
        \hat{d}(m,n) = \sum_{i=0}^{D_{max}} i \times \Phi(-c_{i}(m, n)),
    \end{split}
    \label{eq::1}
\end{equation}
where $\hat{d}(m,n)$ is the predicted disparity at the pixel $(m,n)$, $c_{i}(m,n)$ is the cost corresponding to the $i$-$th$ disparity of the pixel $(m,n)$, $D_{max}$ is the largest disparity and $\Phi(\cdot)$ is the softmax operator. Then if MSE is taken as the regression loss, the loss function is 
\begin{equation}
    \begin{split}
        \frac{1}{MN}\sum_{m,n}^{M,N}(\hat{d}(m,n)-d_{gt}(m,n))_{2}^{2},
    \end{split}
    \label{eq::2}
\end{equation}
where $M,N$ is the height and width of the disparity map, $d_{gt}(m,n)$ is the groundtruth disparity of pixel $(m,n)$. 

$\hat{d}$ is not only a good approximation of $\mathop{\arg\min}\limits_{\small{0 \le i \le D_{max}}}i$, but also can take subpixel information into consideration.  However, just using trivial regression loss is not enough to regularize the cost volume, which is a very essential step in disparity computation. There are still two potential risks. One is that the regression loss can not penalize the multi-modal distributed cost volume, the other is that regression loss may neglect the coherence in the cost volume: costs in a neighbor vary moderately. As for these two risks, GCNet explained ambiguously that the network itself may serve as a regularization to address these problems. Unfortunately, we can still witness many cases that the above two risks occur. 

Firstly, we aim to find a candidate loss to penalize the multi-modal distributed cost volume. The extreme unimodal situation is that the softmax distribution is Dirac delta, and one can easily get $\hat{d}$ equals $\mathop{\arg\min}\limits_{\small{0 \le i \le D_{max}}}i$ if distribution is Dirac delta. From this perspective, a natural solution is adopting one-hot cross entropy widely used in CNN classification as follows:
\begin{equation}
    \begin{split}
        -\sum_{i=0}^{D_{max}}I(i==d_{gt}(m,n))\log \Phi(-c_{i}(m,n)),
    \end{split}
    \label{eq::3}
\end{equation}
where $I(\cdot)$ is the 0-1 indicator function. 

From (\ref{eq::3}), we can see that one-hot cross entropy makes the cost volume put all its mass to the groundtruth pixel. However, the one-hot cross entropy may neglect important subpixel information because of its total concentration. Also, because the softmax probability is fully supported, one-hot groundtruth cannot deal with the coherence problem. 

To mitigate the over-concentration of one-hot cross entropy, loss function of Content-CNN offers a valuable inspiration. In Content-CNN, 3-pixel-hot cross entropy was proposed as follows:
\begin{equation}
    \begin{split}
        -\sum_{i=0}^{D_{max}}\sum_{j=0}^{2}\lambda_{j}I(|i-d_{gt}(m,n)|==j)\log\Phi(-c_{i}(m,n)),
    \end{split}
    \label{eq::4}
\end{equation}
where $\lambda_{i}$ is the loss weight and when $0 \leq i \leq j  \leq 2$, $\lambda_{i} \ge \lambda_{j} \ge 0$. 

The authors explained that the reason for taking only 3 pixels is that $>3$px is an important metric for evaluating stereo algorithms. In (\ref{eq::4}), several adjacent pixels around the groundtruth is distributed unimodally, aiming to preserve the coherence in the local area.  However, only taking several pixels into account may be a sub-optimal solution. Then we aim to find more suitable unimodal distribution for better characterizing specific properties. 

To find the optimal label distribution for the desired property, we need to review general parametric models for regression problem. Given the parametric model $f: \mathcal{X} \to \mathcal{Y}$, where $\mathcal{X}$ is the input space and $\mathcal{Y}$ is the output space. For any paired $x \in \mathcal{X}$ and $y \in \mathcal{Y}$, the regression problem is usually defined as follows:
\begin{equation}
    \begin{split}
        y = f(x) + \epsilon,
    \end{split}
    \label{eq::5}
\end{equation}
where $\epsilon$ is the independent residual error between the prediction and groundtruth. Usually an i.i.d zero-mean Gaussian distribution is adopted for its good properties, that is, $p(y|x) \sim \mathcal{N}(f(x);\sigma^2)$. When $f$ denotes the disparity regression network with the softargmin operator, $f$ can be written as follows:
\begin{equation}
    \begin{split}
        f(x) = \sum_{i=0}^{D_{max}} i\phi_{i},
    \end{split}
    \label{eq::6}
\end{equation}
where $\phi = (\phi_{0}, \dots, \phi_{D_{max}})$ is the desired softmax probability and $0 < f(x) < D_{max}$. Then (5) can be written: 
\begin{equation}
    \begin{split}
        y = \sum_{i=0}^{D_{max}}i\phi_{i} + \epsilon.
    \end{split}
    \label{eq::7}
\end{equation}
Consider the expectation of $y$ given $x$ as follows,
\begin{equation}
    \begin{split}
        \sum_{i=0}^{D_{max}}i \phi_{i} = \int_{A}y \frac{1}{\sqrt{2 \pi} \sigma} e^{-\frac{(y-f(x))^2}{2 \sigma^2}}dy \\+ \int_{A^c}y \frac{1}{\sqrt{2 \pi} \sigma} e^{-\frac{(y-f(x))^2}{2 \sigma^2}}dy,
    \end{split}
    \label{eq::8}
\end{equation}
where $A:= [0, D_{max}]$ and $A^c$ is its complement. Because possible disparity is constrained in interval $A$, we need to eliminate the effects of $A^{c}$. Here we utilize the limitation trick. For $\displaystyle \lim_{\sigma \to 0} \int_{A^c}y \frac{1}{\sqrt{2 \pi} \sigma} e^{-\frac{(y-f(x))^2}{2 \sigma^2}}dy = 0$, that is, the optimal $\phi$ should satisfy
\begin{equation}
    \begin{split}
        \displaystyle \lim_{\sigma \to 0} |\sum_{i=0}^{D_{max}}i \phi_{i} - \int_{A}y \frac{1}{\sqrt{2 \pi} \sigma} e^{-\frac{(y-f(x))^2}{2 \sigma^2}}dy| = 0.
    \end{split}
    \label{eq::9}
\end{equation}

Obviously , if we take $\sum_{i=1}^{D_{max}}i \phi_{i}$ as the Euler discretization of integration $\int_{A}y \frac{1}{\sqrt{2 \pi} \sigma} e^{-\frac{(y-f(x))^2}{2 \sigma^2}}dy$, then (\ref{eq::9}) holds. So under the i.i.d Gaussian observation noise assumption widely used in regression models, given the groundtruth disparity $d_{gt}$, one optimal form of the discrete label distribution $\phi$ is 
\begin{equation}
    \begin{split}
        \phi_{i} = \frac{e^{-\frac{(i-d_{gt})^2}{2\sigma^2}}}{\sum_{j=1}^{D_{max}}e^{-\frac{(j-d_{gt})^2}{2\sigma^2}}}=\Phi(-\frac{(i-d_{gt})^2}{2\sigma^2}). 
    \end{split}
     \label{eq::10}
\end{equation}

Take (\ref{eq::10}) into the cross entropy, the desired loss term is:
\begin{equation}
    \begin{split}
        -\sum_{i=1}^{D_{max}}\Phi(-\frac{(i-d_{gt})^2}{2\sigma^2})\log(\Phi(-c_{i}(m,n))).
    \end{split}
    \label{eq::11}
\end{equation}
From the analysis above, the desired discrete label distribution is determined by the noise distribution, so we call the loss in (\ref{eq::11}) the Gaussian noise-sampling cross entropy loss. Additionally, the i.i.d noise term is often assumed to be Laplacian one for sparse purpose, which is corresponding to MAE. In such cases, we can get the Laplacian noise-sampling cross entropy loss as follows:
\begin{equation}
    \begin{split}
        -\sum_{i=1}^{D_{max}} \Phi(-\frac{|i-d_{gt}|}{\lambda})\log(\Phi(-c_{i}(m,n))).
    \end{split}
    \label{eq::12}
\end{equation}
The proposed noise-sample cross entropy losses $\mathcal{L}_{noise}$ such as (\ref{eq::11}) and (\ref{eq::12}) aim to regularize the cost volume to be unimodal and coherent, and its form should be consistent with the regression loss. Given the softargmin regression loss $\mathcal{L}_{regress}$ such as (\ref{eq::2}), the final loss contains two parts:
\begin{equation}
    \begin{split}
        \mathcal{L} = \mathcal{L}_{regression} + \mu \mathcal{L}_{noise}, 
    \end{split}
    \label{eq::13}
\end{equation}
where $\mu$ is a hyper-parameter and is set between $(0, 1)$. The sketch of the whole framework is depicted in Fig.\ref{fig:network}.

\section{Experimental Results} \label{sec:experiment}

\subsection{Dataset and Implementation Details}
We preformed our method on the KITTI 2015 dataset. The KITTI 2015 is a dataset of the real-world traffic situation scenario images captured by a driving vehicle. The image size is 376 $\times$ 1240. The dataset~\cite{geiger2013vision} provided 200 pairs of  stereo images with sparse groundtruth disparities for training and 200 pairs of images for testing. 

We trained our network at the learning rate of 0.00001 for the first 200 epochs and 0.000005 for the rest 300 epochs and obtained our best model. The batch size was set to 2 for the training process. The loss weight $\mu$ in noise-sampling cross entropy loss is set as 0.05 in all relevant experiments. 



\begin{figure}[t]
\begin{minipage}[b]{0.48\linewidth}
  \centering
  \centerline{\includegraphics[width=4cm]{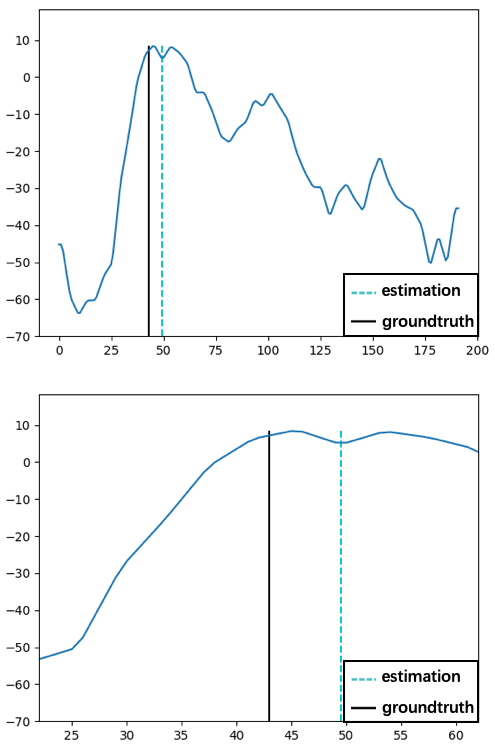}}
  \centerline{(a) \footnotesize{PSMNet} }\medskip
\end{minipage}
\hfill
\begin{minipage}[b]{0.48\linewidth}
  \centering
   \centerline{\includegraphics[width=4cm]{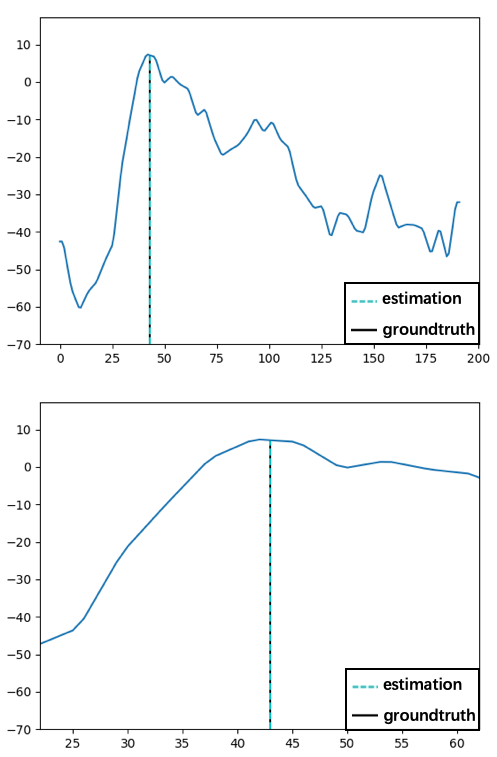}}%
  \centerline{(b) \scriptsize{PSMNet with Noise-sampling loss} }\medskip
\end{minipage}
\caption{The figure shows two examples of the disparity estimation distributions over the cost volume's disparity dimension, which respectively presents two different pixels. The first row of (a) is obtained by PSMNet~\cite{chang2018pyramid}
, while (b) is the softmax version of cost volume produced by PSMNet trained with Laplacian noise-sampling cross entropy loss. To get more detailed information, we zoom into the curves in the second row of each. The dashed line denotes the disparity estimation, and the solid line shows the groundtruth.}
\label{fig:curves}
\end{figure}

\subsection{Ablation Studies on Noise-sampling Cross Entropy Losses}

In this subsection, we experimentally demenstrate the effectiveness of the proposed noise-sampling loss. Here we train the PSMNet~\cite{chang2018pyramid} with four kinds of loss under the same hyper-parameter setting: $\mathcal{L}_{1}$, $\mathcal{L}_{1}+$ neighbor centered cross entropy loss in (\ref{eq::4}), $\mathcal{L}_{2}+$ Gaussian noise-sampling cross entropy loss in (\ref{eq::11}) and $\mathcal{L}_{1}+$ Laplacian noise-sampling cross entropy loss in (\ref{eq::12}), and show their validation error on KITTI 2015 in Table.\ref{table::1} and disparity estimation distributions produced by $\mathcal{L}_{1}$ and $\mathcal{L}_{1}+$ Laplacian noise-sampling cross entropy loss in Fig.\ref{fig:curves}. 

From Fig.\ref{fig:curves}, we can see the cost volume regularized by the Laplacian noise-sampling cross entropy loss is obviously more unimodal and coherent than the one produced by regression loss only. From Table \ref{table::1} we can see that two noise-sampling cross entropy losses both perform better than the naive neighbor centered  cross entropy loss used in \cite{luo2016efficient}, validating the explanation in Section \ref{sec: methods} that the cost volume regularizers taking the regression model into account are much more effective. 

\subsection{Experimental Comparisons with Representative Methods}
\begin{table}[t]
\begin{center}

\caption{Result comparisons among different losses. Common regression loss along with the proposed noise-sampling losses ($\mathcal{L}_{2}+$Gaussian \& $\mathcal{L}_{1}+$Laplacian) can achieve better performance.}\label{table::1}

\resizebox{84mm}{4.5mm}{
\begin{tabular}{|c|c|c|c|c|}
    \hline
    { }&{\bf $\mathcal{L}_{1}$}&{\bf $\mathcal{L}_{1}$ + neighbor} & {\bf $\mathcal{L}_{2}$ + Gaussian} & {\bf $\mathcal{L}_{1}$ + Laplacian} \\
    \hline
    {\bf KITTI 2015 val error(\%)} & 0.830 &  0.831  & 0.826  & 0.824\\
    \hline
\end{tabular}
}
\end{center}
\end{table}


\begin{table}[]
\setlength{\abovecaptionskip}{0.cm}
 
\setlength{\belowcaptionskip}{0.5cm}
\begin{center}

\caption{Results on the KITTI 2015 dataset, our results are submitted on November 10, 2019.}\label{table::2}

\resizebox{84mm}{14mm}{
\begin{tabular}{|c|c|c|c|c|c|c|}

    \hline
    \multirow{2}{*}{\bf Method}&
    \multicolumn{3}{c|}{\bf All(\%)}&
    \multicolumn{3}{c|}{\bf Noc(\%)}\cr\cline{2-7}
    &{\bf D1-bg} & {\bf D1-fg} & {\bf D1-all}&{\bf D1-bg} & {\bf D1-fg} & {\bf D1-all}\\ 
    \hline
    {\bf PSM~\cite{chang2018pyramid}} & 1.86  & 4.62  & 2.32  & 1.71 & 4.31 & 2.14 \\
    \hline
    {\bf GC~\cite{kendall2017end}}  & 2.21 & 6.16 & 2.87 & 2.02 & 5.58 & 2.61 \\%
    \hline
    {\bf SGM~\cite{seki2017sgm}}& 2.66  & 8.64  & 3.66 & 2.23  & 7.44  & 3.09 \\%
    \hline
    {\bf CFP~\cite{zhu2019multi}}& 1.90  & 4.39  & 2.31  & 1.73  & 3.92  & 2.09 \\%
    \hline
    {\bf ours}& 1.86  & 4.35  & 2.27 & 1.71  & 4.08  & 2.10 \\
    \hline
\end{tabular}
}
\end{center}
\end{table}


For authoritative comparisons, we utilize the disparity maps on the KITTI 2015 test dataset of the 200 image pairs and upload the results to the KITTI evaluation server. The uploaded results are reported of $\bf 2.27\%$ 3-pixel error, outperforming the PSMNet~\cite{chang2018pyramid} which is the benchmark network our work based on. According to the online leaderborad, our work surpassed some prior studies as shown in Table \ref{table::2}. As an illustration, the ``All'' columns present the error estimation over all pixels. On the contrary, the ``Noc'' denotes the error only over the non-occluded areas. The ``D1'' means the percentage of stereo disparity outliers in first frame, when the ``bg'', ``fg'', ``all'' present the percentage of outliers averaged only over background, foreground and all ground truth regions.

For qualitative comparisons,  Fig.\ref{fig:res} shows some output image pairs including the result images and corresponding error maps of our noise-sampling PSMNet(NS-PSMNet), PSMNet~\cite{chang2018pyramid}, Content-CNN~\cite{luo2016efficient} and GC-Net~\cite{kendall2017end} given by the KITTI evaluation server. It can be easily figured out that our model yields smoother details over some ill-posed regions in Fig.\ref{fig:res}.

\section{Conclusion} \label{sec:concl}
In this paper, we propose noise-sampling cross entropy loss to regularize the cost volume in recent deep learning based stereo matching algorithms, to acquire unimodal and coherent cost volume. Relevant explanation and experiments demonstrate that with the aid of the proposed loss, the matching results of the current high-performance deep disparity regression models can be further improved. 

\bibliographystyle{IEEEbib}
\bibliography{refs}

\end{document}